%% file: 00_RobustTrend_main.tex
\documentclass{article}
\usepackage{ijcai19}

\usepackage{float}
\usepackage{multirow}
\usepackage{times}
\usepackage{soul}
\usepackage{url}
\usepackage[hidelinks]{hyperref}
\usepackage[utf8]{inputenc}
\usepackage[small]{caption}
\usepackage{graphicx}
\usepackage{amsmath}
\usepackage{booktabs}
\usepackage{algorithm}
\usepackage{algorithmic}
\usepackage{subfigure}

\title{RobustTrend: A Huber Loss with a Combined First and Second Order Difference Regularization for Time Series Trend Filtering}

\author{Qingsong Wen\and Jingkun Gao\and Xiaomin Song\and Liang Sun \And Jian Tan
\affiliations
    Machine Intelligence Technology, Alibaba Group, 
Bellevue, Washington 98004, USA\\ \emails
    \{qingsong.wen, jingkun.g, xiaomin.song, liang.sun, j.tan\}@alibaba-inc.com
}

\begin{document}

\maketitle

\input{0_abstract.tex}

\input{1_introduction.tex}

\input{2_related_work.tex}

\input{3_methodology.tex}

\input{4_experiment.tex}


\input{5_conclusion.tex}

\bibliographystyle{named}

\bibliography{6_robustTrend_bibFile}

\end{document}

%% file: 0_abstract.tex
\begin{abstract}
Extracting the underlying trend signal is a crucial step to facilitate time series analysis like forecasting and anomaly detection. Besides noise signal, time series can contain not only outliers but also abrupt trend changes in real-world scenarios. To deal with these challenges, we propose a robust trend filtering algorithm based on robust statistics and sparse learning. Specifically, we adopt the Huber loss to suppress outliers, and utilize a combination of the first order and second order difference on the trend component as regularization to capture both slow and abrupt trend changes. Furthermore, an efficient method is designed to solve the proposed robust trend filtering based on majorization minimization (MM) and alternative direction method of multipliers (ADMM). We compared our proposed robust trend filter with other nine state-of-the-art trend filtering algorithms on both synthetic and real-world datasets. The experiments demonstrate that our algorithm outperforms existing methods.
\end{abstract}

%% file: 1_introduction.tex
\section{Introduction}\label{sec:intro}






The explosion of time series data generated by the Internet of Things (IoT) and many other sources has made the time series analytics more challenging. Trend filtering, which estimates the underlying trend from the time series, is an important task arising in a variety of real-world applications~\cite{L1trend_kim2009ell,craigmile2006wavelet}. Specifically, we assume the time series consists of the trend component and a more rapidly changing random component. The goal of trend filtering is to extract the trend component accurately and efficiently. The extracted trend component can reveal certain traits of a time series and is crucial in many applications such as time series forecasting and anomaly detection.

Many algorithms have been proposed for trend filtering, including moving average smoother, kernel smoothing, smoothing splines etc~\cite{wu2007trend,wu2009ensemble,fried2018package,afanasyev2016long}. Among them, one of the most classical and widely used trend filters is the Hodrick-Prescott (H-P) filtering~\cite{hp_hodrick1997} and its variants. In the H-P trend filtering, the trend component is obtained by solving an optimization problem which minimizes the weighted sum of the residual size and the smoothness of the trend component. On the one hand, the extracted trend component is expected to close to the original time series, but meanwhile the smoothness constraint is imposed. Specifically, the sum-of-squares loss function is used to measure the difference between the trend and the original time series, and the smoothness of the trend component is measured using the second order difference of the trend. The drawback of H-P filtering is that it is not robust to outliers. Based on the H-P filtering, the $\ell_1$ trend filtering~\cite{L1trend_kim2009ell} is proposed, which basically substitutes the $\ell_1$-norm for the $\ell_2$-norm in the regularization term in H-P trend filtering. Note that the $\ell_1$ trend filtering assumes that the underlying trend is piecewise linear, and the kinks, knots, or changes in slope of the
estimated trend can be interpreted as abrupt changes or events. 

Although the sum-of-squares loss function in both the H-P trend filtering and the $\ell_1$ trend filtering enjoys its popularity, it cannot handle heavy tailed error or outliers. One approach to deal with these situations is the least absolute deviation (LAD) loss~\cite{lad:lasso} as it mitigates the loss incurred by big errors. However, the LAD loss is not adapted for small errors since it penalizes strongly on the small residuals. In the extreme case where the error is not heavy tailed or there is no outliers, it is less effective than the sum-of-squares loss function. In practice, generally we expect the loss function can adaptively handle these cases. The Huber loss~\cite{huber2011robust} is a combination of the sum-of-squares loss and the LAD loss, which is quadratic on small errors but grows linearly for large values of errors. As a result, the Huber loss is not only more robust against outliers but also more adaptive for different types of data. 

In this paper, besides adopting Huber loss, we propose to use the combination of the first order and the second order difference regularization of the trend component in trend filtering. The regularization based on the first order difference encourages to preserve abrupt trend change, but meanwhile it tends to introduce the staircases for the extracted trend component when slow trend change exists. One way to reducing the staircasing effect is the introduction of higher order difference in the regularizier. In fact, the second order difference can effectively eliminates the artifacts (e.g., staircasing) introduced by the first regularization without smoothing the time series strongly. Also, it is proved recently that the introduction of the second order difference can achieve comparable results to higher order difference ~\cite{Papafitsoros2014}. Thus, we consider the combination of the first and second order difference for the best trade-off.  

To sum up, in this paper we propose a novel trend filtering method based on the Huber loss and the combination of the first and second order dynamics of the trend component. The resulting optimization problem is solved via an efficient MM-ADMM algorithm. The complexity of the Huber loss makes it challenging to directly apply the ADMM framework~\cite{boyd2011distributed}. Inspired by the majorization minimization~\cite{Sun:majorization:mini}, we minimize an upper bound of the Huber loss in the ADMM framework. In the experiments, we compare ten different trend filtering algorithms and results on both synthetic and real-world data sets demonstrate the good performance of our trend filtering algorithm. To the best of our knowledge, our empirical comparisons is one of the most extensive and comprehensive studies which includes common popular trend filtering methods, including H-P trend filter, $\ell_1$ trend filter, {TV denoising filter}~\cite{chan2001digital}, {mixed trend filter}~\cite{tibshirani2014adaptive}, {wavelet trend filter}~\cite{craigmile2006wavelet}, {repeated median filter}~\cite{siegel1982robust}, {robfilter}~\cite{fried2004robust}, {EMD trend filter}~\cite{cmp_l1hpemp}, and so on.




The rest of the paper is organized as follows. In Section 2, we briefly review the trend filtering and some related work. In Section 3, We introduce our robust trend filter and our efficient algorithm. We present the empirical studies of the proposed method in comparison with other state-of-the-art algorithms in Section 4, and conclude the paper in Section 5. 

%% file: 2_related_work.tex
\section{Related Work}\label{sec:related}

There are various trend filtering methods proposed in the literature. As discussed in the previous section, the H-P trend filtering~\cite{hp_hodrick1997} and $\ell_1$ trend filtering~\cite{L1trend_kim2009ell} extract the trend component by minimizing the weighted sum of the residual size and the smoothness of the trend component using the second order difference operator. The TV denoising filtering ~\cite{chan2001digital} adopts the same loss function as the H-P and $\ell_1$ trend filtering, but it utilizes the first order difference as the regularization to deal with abrupt trend changes. As a result, the TV denoising filtering can handle the abrupt trend change successfully, but it tends to introduce staircases for slow trend changes. In contrast, the H-P and $\ell_1$ trend filtering generally prone to trend delay for abrupt trend changes. The mixed trend filtering~\cite{tibshirani2014adaptive} adopts two difference orders as regularization. It is assumed that the observed time series were drawn from an underlying function possessing different orders of piecewise polynomial smoothness. However, it is difficult to test this assumption adaptively. Thus, the selection of the two order remains a challenging problem, and limits its usage in practice. Besides, due to the sum-of-squares loss, all the aforementioned trend filtering methods cannot handle heavy tailed errors and outliers.

In~\cite{fried2004robust,fried2018package}, a robust trend filtering called robfilter is proposed. In robfilter, a robust regression functional for local approximation of the trend is used in a moving window. To further improve the robustness of robfilter, outliers in the original time series is removed by trimming or winsorization based on robust scale estimators. 
Empirical mode decomposition (EMD)~\cite{wu2007trend} is a popular method for analyzing non-stationary time series. With iterative process, it finds local maxima and minima, and then interpolates them into upper and lower envelopes. The mean of the envelopes yields local trend, while the residual must comply two requirements to become an intrinsic mode function (IMF). This method assumes that the trend is ``slowly varying" and the selection of ``slowest" IMF sums up to be the trend. Since the local maxima and minima are sensitive to noise and outliers, the conventional EMD are not robust. Another variant, Ensemble EMD (EEMD), is a noise-assisted method. It is an ensemble of white noise-added data and treats mean of their output as the final true result. It resolves the mode mixing issue in original EMD but still suffer from outliers.
The wavelet analysis~\cite{afanasyev2016long} is a transformation of the original time series into two types of coefficients, wavelet coefficients and scaling coefficients. Note that the wavelet coefficients are related to changes of average over specific scales, whereas scaling coefficients can be associated with averages on a specified scale. Since the scale associated with scaling coefficients is usually fairly large,  the general idea behind the wavelet trend filter~\cite{craigmile2006wavelet} is to associate the scaling coefficients with the trend, and the wavelet coefficients with the noise component. The benefits of trend filtering via wavelet transformation is that it can capture the abrupt change of trend, but how to choose orthogonal basis still remains a challenge. 

In this paper, we focus on trend filtering for time series without distinguishing periodic and trend components. When the seasonality/periodic and trend components need to be separated, the seasonal-trend decomposition method can be utilized, such as STL~\cite{STL_cleveland1990stl}, RobustSTL~\cite{RobustSTL_wen2018robuststl}, and TBATS~\cite{TBATS_de2011forecasting}.

%% file: 3_methodology.tex
\section{Proposed Robust Trend Filter}\label{sec:method}

\subsection{Model Overview} \label{sec:sysmodel} 
We consider the time series of length $N$ as $\mathbf{y}=[y_0, y_1, \cdots, y_{N-1}]^T$, which can be decomposed into trend and residual components ~\cite{alexandrov2012review}
\begin{equation}\label{eq:whole_model} 
y_t = \tau_t + r_t \quad\text{or}\quad \mathbf{y} = \boldsymbol{\tau} + \mathbf{r},
\end{equation}
where $\boldsymbol{\tau}=[\tau_0, \tau_1, \cdots, \tau_{N-1}]^T$ denotes the trend component, and $\mathbf{r} = [r_0, r_1, \cdots, r_{N-1}]^T$ denotes the residual component. Note that in this paper we consider the time series faced in real-world applications. Therefore, the trend component may contains both slow trend change and abrupt trend change, while the residual component may contains both noise and outliers.

\subsection{Preliminary} \label{sec:Preliminary}
In the H-P trend filtering~\cite{hp_hodrick1997}, where the trend estimation $\tau_{t}$ is chosen to minimize the objective function as follows
\begin{equation}\label{eq:hp_trend} 
\frac{1}{2} \sum_{t=0}^{N-1} ( y_t - \tau_{t} )^2
+ \lambda \sum_{t=1}^{N-2} ( \tau_{t-1} - 2\tau_{t} + \tau_{t+1} )^2,
\end{equation}
where the loss function (first term) measures the size of the residual signal after trend extraction while the regularization (second term) measures the smoothness of the extracted trend, and the regularization parameter $\lambda>0$ controls the trade-off between them. The Eq.~\eqref{eq:hp_trend} can be equivalently formulated in the matrix form as
\begin{equation}\label{eq:hp_trend_matrix} 
\frac{1}{2} || \mathbf{y} - \boldsymbol{\tau} ||_2^2 + \lambda || \mathbf{D}^{(2)}\boldsymbol{\tau} ||_2,
\end{equation}
where $\mathbf{D}^{(2)}$ is the second-order difference matrix, and its definition is given in Eq.~\eqref{eq:Dk} below. 

By replacing the $\ell_2$-norm with $\ell_1$-norm in the regularization term of H-P trend filtering, we obtain the $\ell_1$ trend filtering~\cite{L1trend_kim2009ell} as follows: 
\begin{equation}\label{eq:L1_trend_matrix} 
\frac{1}{2} || \mathbf{y} - \boldsymbol{\tau} ||_2^2 + \lambda || \mathbf{D}^{(2)}\boldsymbol{\tau} ||_1.
\end{equation}
The $\ell_1$ regularization in Eq.~\eqref{eq:L1_trend_matrix} promotes sparsity, which indicates $|| \mathbf{D}^{(2)}\boldsymbol{\tau} ||_1$ would have many zero elements. As a result, the $\ell_1$ trend filter leads to a piecewise linear trend component. 

As an extension of Eq.~\eqref{eq:L1_trend_matrix}, piecewise polynomial trend signal can be obtained by using high-order difference operator in the regularization term~\cite{L1trend_kim2009ell}, i.e.,
\begin{equation}\label{eq:L1_trend_kth} 
\frac{1}{2} || \mathbf{y} - \boldsymbol{\tau} ||_2^2 + \lambda || \mathbf{D}^{(k+1)}\boldsymbol{\tau} ||_1,
\end{equation}
where $\mathbf{D}^{(k+1)} \in \mathcal{R}^{(n-k-1)\times n}$ is the discrete difference operator of order $k+1$, which can be defined recursively as ${D}^{(k+1)} = {D}^{(1)} \cdot {D}^{(k)}$ and the
${D}^{(1)}$ and ${D}^{(2)}$ are the first-order and second-order difference, i.e.,
\begin{equation} \label{eq:Dk}
\resizebox{1.0\hsize}{!}{$
\mathbf{D}^{(1)} =  
\begin{bmatrix}
1 & -1\\
  & 1& -1\\
& &\ddots\\
&&1&-1
\end{bmatrix},
~\mathbf{D}^{(2)} = 
\begin{bmatrix}
1 & -2 &1&\\
&1&-2&1&\\
&&&\ddots\\
&&&1&-2&1
\end{bmatrix}.
$}
\end{equation}
When $k=0$ in Eq.~\eqref{eq:L1_trend_kth}, it corresponds to the total variation (TV) filter~\cite{chan2001digital}, which is suitable when the underlying trend signal is approximately piecewise constant.



\subsection{Robust Trend Filtering with Sparse Model} \label{sec:RobustTrend} 

All these trend filtering methods discussed in Section~\ref{sec:Preliminary} adopt the sum-of-squares loss, which is appropriate when the residual part is near to Gaussian noise. However, in real-world applications, the residual often exhibits long-tailed distribution due to outliers. In this paper, we adopt the Huber loss from robust statistics~\cite{huber2011robust} to deal with possible long-tailed distributed residual. As discussed in Section~\ref{sec:intro}, Huber loss combines the advantages of the sum-of-squares loss and the LAD loss adaptively.


In order to simultaneously capture abrupt and slow trend changes, we consider two sparse regularization terms like the fussed LASSO~\cite{tibshirani2005sparsity}. Specifically, we utilize first and second order difference operators in the regularization to deal with abrupt and slow trend changes, respectively. Since it is proved that the combination of the first and second order difference regularization can achieve comparable results to higher order difference regularization~\cite{Papafitsoros2014}, in this paper we do not consider other higher order difference operators. 

Based on the above discussion, we propose a Huber loss with a combined first and second order difference regularization as a robust trend filtering for time series (named RobustTrend in this paper), i.e.,
\begin{equation}\label{eq:robust_trend} 
g_{\gamma} (\mathbf{y} - \boldsymbol{\tau}) 
+ \lambda_1 || \mathbf{D}^{(1)}\boldsymbol{\tau} ||_1
+ \lambda_2 || \mathbf{D}^{(2)}\boldsymbol{\tau} ||_1
\end{equation}
where ${D}^{(1)}$, ${D}^{(2)}$ are the first-order and second-order difference matrix, respectively. And $g_{\gamma} (\mathbf{x})=\sum_i g_{\gamma} (x_i)$ is the summation of elementwise Huber loss function with each element as
\begin{equation}\label{eq:huber_loss} 
 g_{\gamma} (x_i) =
\begin{cases} 
\frac{1}{2} x_i^2,       & |x_i| \leq \gamma\\
\gamma|x_i|- \frac{1}{2} \gamma^2, & |x_i| > \gamma
\end{cases}
\end{equation}
and its derivative can be obtained as
\begin{equation}\label{eq:huber_derivative} 
g_{\gamma}' (x_i) =
\begin{cases} 
x_i,   & |x_i| \leq \gamma \\
\gamma \text{sgn}(x_i),   & |x_i| > \gamma
\end{cases}
\end{equation}
where $\text{sgn}$ is the sign function.

\subsection{Efficient MM-ADMM Algorithm} \label{sec:mm-admm}
In this section, we develop a specialized ADMM algorithm to solve the minimization problem of RobustTrend in Eq.~\eqref{eq:robust_trend}. ADMM~\cite{boyd2011distributed} is a powerful framework in optimization and also utilized in trend filtering with sum-of-square loss~\cite{ramdas2016fast}. However, the standard ADMM is not efficient in the RobustTrend filter since no exact solution exists in the updating step with Huber loss (will be shown later). One may transform Eq.~\eqref{eq:robust_trend} into a series of ordinary fused lasso problems as in~\cite{polson2016mixtures}. But it is still inefficient since each ordinary fused lasso problem needs to be solved iteratively as well.
To alleviate this problem, we utilize the majorization-minimization (MM) framework~\cite{Sun:majorization:mini} to approximate the Huber loss function in the ADMM updating step to obtain a closed form for efficient computation.

To simplify the ADMM formulation for Eq. \eqref{eq:robust_trend}, we rewrite it in an equivalent form as
\begin{equation}\label{eq:robust_trend_opt} 
\begin{matrix}
\min & 
g_{\gamma} (\mathbf{y} - \boldsymbol{\tau}) 
+ || \mathbf{z} ||_1 \\
\textrm{s.t.} & 
\mathbf{D}\boldsymbol{\tau} = \mathbf{z} 
\end{matrix}
\end{equation}
where 
\begin{equation}\label{eq:matrix_D} 
\mathbf{D}= 
\begin{bmatrix}
\lambda_1 \mathbf{D}^{(1)}\\
\lambda_2 \mathbf{D}^{(2)}\\
\end{bmatrix}.
\end{equation}
Then we can obtain the augmented Lagrangian as
\begin{equation}\label{eq:Lagrangian} \notag
L_{\rho}(\boldsymbol{\tau},\mathbf{z},\mathbf{v}) = 
g_{\gamma} (\mathbf{y} - \boldsymbol{\tau}) + || \mathbf{z} ||_1 +
\mathbf{v}^T (\mathbf{D}\boldsymbol{\tau} - \mathbf{z}) + \frac{\rho}{2} ||\mathbf{D}\boldsymbol{\tau} - \mathbf{z}||_2^2
\end{equation}
where $\mathbf{v}$ is the dual variable, and $\rho$ is the penalty parameter.
Following the ADMM procedure in~\cite{boyd2011distributed}, we can obtain the updating steps as
\begin{align} 
\boldsymbol{\tau}^{k+1} &= \underset{\boldsymbol{\tau}}{\arg\min}
\left(g_{\gamma} (\mathbf{y} - \boldsymbol{\tau}) + \frac{\rho}{2} || \mathbf{D}\boldsymbol{\tau}  - \mathbf{z}^k + \mathbf{u}^k  ||_2^2    \right)  \label{eq:admm_1}\\
\mathbf{z}^{k+1} &= \underset{\mathbf{z}}{\arg\min}
\left(||\mathbf{z}||_1 + \frac{\rho}{2} || \mathbf{D}\boldsymbol{\tau}^{k+1}  - \mathbf{z} + \mathbf{u}^k  ||_2^2    \right)\label{eq:admm_2}\\ 
\mathbf{u}^{k+1} &= \mathbf{u}^{k} + \mathbf{D}\boldsymbol{\tau}^{k+1} - \mathbf{z}^{k+1}\label{eq:admm_3} 
\end{align}
where  $\mathbf{u}=(1/\rho)\mathbf{v}$ is the scaled dual variable to make the formulation more convenient. The above $\mathbf{z}-$minimization step can be efficiently solved by soft thresholding as 
\begin{equation}\label{eq:z_minimization} 
\mathbf{z}^{k+1} = {S}_{\rho}( \mathbf{D}\boldsymbol{\tau}^{k+1}  + \mathbf{u}^k)
\end{equation}
where the soft thresholding operator ${S}_{\rho}(a)$ is defined as
\begin{equation}\label{eq:soft_thresholding} 
{S}_{\rho}(a) =
\begin{cases} 
0,       & |a| \leq \rho\\
a - \rho \text{sgn}(a), & |a| > \rho
\end{cases}
\end{equation}

The $\boldsymbol{\tau}-$minimization step in Eq.~\eqref{eq:admm_1} is nontrivial since no exact solution exists. One may use iterative algorithms, like standard gradient methods or conjugate gradient methods, to solve the $\boldsymbol{\tau}-$minimization step since it is convex. However, this would increase the overall computation since the ADMM itself is an iterative algorithm. Inspired by the work of~\cite{eckstein1992douglas} that the ADMM can still converge when the minimization steps are carried out approximately, we utilize one-iteration MM framework like ~\cite{pham2013efficient} to efficiently solve the $\boldsymbol{\tau}-$minimization step in Eq.~\eqref{eq:admm_1} with closed form solution. Specifically, we adopt the proved sharpest quadratic majorization for Huber loss as in ~\cite{de2009sharp} to minimize the difference between the exact and approximate solution, i.e.,
\begin{align}\label{eq:huber_mm}  \notag
g_{\gamma} (x_i) \leq \eta_{\gamma} (x_i | x_i^k) 
&= \frac{g_{\gamma}'(x_i^k)}{2x_i^k} (x_i^2 -(x_i^k)^2) + g_{\gamma} (x_i^k) \\\notag
&= \frac{g_{\gamma}'(x_i^k)}{2x_i^k} x_i^2 + C,
\end{align}
where $\eta_{\gamma} (x_i | x_i^k)$ denotes the majorization of element $x_i$'s Huber loss and $x_i^k$ is the solution of $x_i$ at ADMM's $k$th iteration. Then, the summation of Huber loss can be obtained as
\begin{equation}\label{eq:huber_mm_matrix} 
g_{\gamma} (\mathbf{x}) \leq \eta_{\gamma} ( \mathbf{x} | \mathbf{x}^k) = \sum_i \eta_{\gamma} (x_i | x_i^k) 
= \frac{1}{2} \mathbf{x}^T \mathbf{A} \mathbf{x} + C
\end{equation}
where $C$ represents a constant number and $\mathbf{A}$ is a diagonal matrix as   
\begin{equation}\label{eq:huber_mm_A} 
\mathbf{A} = \text{diag}\left(g_{\gamma}' (x_i^k)\right) \text{diag}^{-1}\left(g_{\gamma} (x_i^k)\right).
\end{equation}
Next, let define $\mathbf{x} = \mathbf{y} - \boldsymbol{\tau}$ and replace the Huber loss in Eq.~\eqref{eq:admm_1} with the quadratic majorization in Eq.~\eqref{eq:huber_mm_matrix}, we can obtain an approximate $\boldsymbol{\tau}-$minimization step with one-iteration MM algorithm as
\begin{align} 
\mathbf{x}^{k+1} &= \underset{\mathbf{x}}{\arg\min}
\left( \frac{1}{2} \mathbf{x}^T \mathbf{A} \mathbf{x} + \frac{\rho}{2} || \mathbf{D}\mathbf{x} - (\mathbf{u}^k - \mathbf{z}^k +  \mathbf{Dy})  ||_2^2    \right)\label{eq:t_min_1}  \\
\boldsymbol{\tau}^{k+1} &= \mathbf{y} - \mathbf{x}^{k+1}\label{eq:t_min_2}
\end{align}
To obtain the closed form of the above approximate $\boldsymbol{\tau}-$minimization step,
we take derivative of Eq.~\eqref{eq:t_min_1}, set it to zero to obtain $\mathbf{x}^{k+1}$, and put $\mathbf{x}^{k+1}$ into Eq.~\eqref{eq:t_min_2} to get the final $\boldsymbol{\tau}-$minimization step as
\begin{equation}\label{eq:final_tau} 
\boldsymbol{\tau}^{k+1} = \mathbf{y} - \rho (\mathbf{A} + \rho  \mathbf{D}^T \mathbf{D} )^{-1} \mathbf{D}^T (\mathbf{u}^k - \mathbf{z}^k +  \mathbf{Dy}).
\end{equation}
Therefore, to obtain the extracted trend signal of the proposed RobustTrend filter, our designed MM-ADMM algorithm sequentially calculates \textbf{Eq.~(21)}, \textbf{Eq.~(15)}, and \textbf{Eq.~(14)} in each ADMM iteration until termination. 


For the termination criteria, besides the maximum iteration number, we also check the primal and dual residuals in each ADMM iteration until they are small enough. Specifically, the values of primal residual $\mathbf{r}^k$ and dual residual $\mathbf{r}^k$ at $k$th iteration are calculated by
\begin{align}\label{eq:primal:residual}
\|\mathbf{r}^k\|_2 &= \|\mathbf{D}\boldsymbol{\tau}^{k} - \mathbf{z}^k\|_2,\\
\|\mathbf{s}^k\|_2 &= \rho \|\mathbf{D}^T (\mathbf{z}^k - \mathbf{z}^{k-1})\|_2.
\end{align}
Then, $\|\mathbf{r}^k\|_2$ and $\|\mathbf{s}^k\|_2$ are checked if they are smaller than the corresponding tolerance thresholds as~\cite{boyd2011distributed}
\begin{align} \notag
\epsilon^{pri} &= \sqrt{2N-3} \epsilon^{abs} + \epsilon^{rel}\max\{\|\mathbf{D}\boldsymbol{\tau}^{k}\|_2, \|\mathbf{z}^{(k)}\|_2 \}, \\\notag
\epsilon^{dual} &= \sqrt{N}\epsilon^{abs} + \epsilon^{rel}\|\rho \mathbf{D}^T \mathbf{u}^{(k)}\|_2,
\end{align}
where $\epsilon^{abs}>0$ is an absolute tolerance and $\epsilon^{rel}>0$ is a relative tolerance. A reasonable choice for the two tolerance is around $10^{-3}$ to $10^{-4}$.

\subsection{Online Extension for Data Streams} 

In order to support streaming data applications, the proposed RobustTrend filter is extended to online mode by applying sliding window scheme like robfilter and repeated median filter in~\cite{fried2004robust,fried2018package}. During each sliding window, the trend signal is extracted using our designed MM-ADMM algorithm described in Section \ref{sec:mm-admm}. To speed up, the $\rho\mathbf{D}^T\mathbf{D}$ and $\mathbf{Dy}$ are computed once and cached for the $\boldsymbol{\tau}-$minimization step in Eq.~\eqref{eq:final_tau}. Furthermore, warm start is adopted in each new sliding window, where the initial values for $\boldsymbol{\tau}$, $\mathbf{z}$, and $\mathbf{u}$ are from previous window's final solution by removing the first element and copying last element once. Since the trend signal usually exhibits slow change, this warm start trick often provides an appropriate approximation to result in fewer ADMM iterations.



%% file: 4_experiment.tex
\section{Experiments and Comparisons}
In this section, we conduct experiments on both synthetic data and real-world data to demonstrate the effectiveness of the proposed RobustTrend filtering algorithm.

\begin{table*}[!htb]
\footnotesize \centering

\begin{tabular}{l|rrrr|rrrr}
\hline
Metric & \multicolumn{4}{c|}{MSE} & \multicolumn{4}{c}{MAE} \\
Outlier ratio &      1\% &      5\% &     10\% &     20\%  &      1\% &      5\% &     10\% &     20\% \\
\hline\hline
H-P filter                  &  0.0126 &  0.0209 &  0.0263 &  0.0502 &  0.0656 &  0.0985 &  0.1158 &  0.1730 \\
$\ell_1$ trend filter             &  0.0102 &  0.0170 &  0.0226 &  0.0469 &  0.0536 &  0.0860 &  0.1029 &  0.1684 \\
TV denoising filter         &  0.0184 &  0.0223 &  0.0277 &  0.0473 &  0.0973 &  0.0983 &  0.1272 &  0.1827 \\
Mixed trend filter          &  0.0070 &  0.0137 &  0.0183 &  0.0451 &  0.0562 &  0.0772 &  0.0963 &  0.1715 \\
Wavelet trend filter        &  0.0167 &  0.0293 &  0.0348 &  0.0648 &  0.0857 &  0.1300 &  0.1400 &  0.1996 \\
robfilter                   &  \textbf{0.0047} &  0.0109 &  0.0080 &  0.0110 &  0.0500 &  0.0569 &  0.0596 &  \textbf{0.0586} \\
Repeated median filter      &  0.0121 &  0.0117 &  0.0129 &  0.0145 &  0.0615 &  0.0599 &  0.0642 &  0.0724 \\
EMD filter                  &  0.0240 &  0.0366 &  0.0401 &  0.0902 &  0.0999 &  0.1410 &  0.1523 &  0.2293 \\
EEMD filter                 &  0.0200 &  0.0286 &  0.0393 &  0.0595 &  0.0917 &  0.1189 &  0.1479 &  0.1850 \\
Proposed RobustTrend filter &  0.0051 &  \textbf{0.0054} &  \textbf{0.0058} &  \textbf{0.0079} &  \textbf{0.0434} &  \textbf{0.0442} &  \textbf{0.0501} &  0.0638 \\
\hline
\end{tabular}
\caption{Performance of the proposed algorithm compared with other trend filter algorithms on synthetic data with different ratios of outliers.}
\label{tab: 1-alg-comparison-synthetic}
\vspace{-0.3cm}

\end{table*}

\subsection{Baseline Algorithms}

\begin{figure}[!htb]
    \centering
    \subfigure[Synthetic Data]{
        \includegraphics[width=0.9\linewidth]{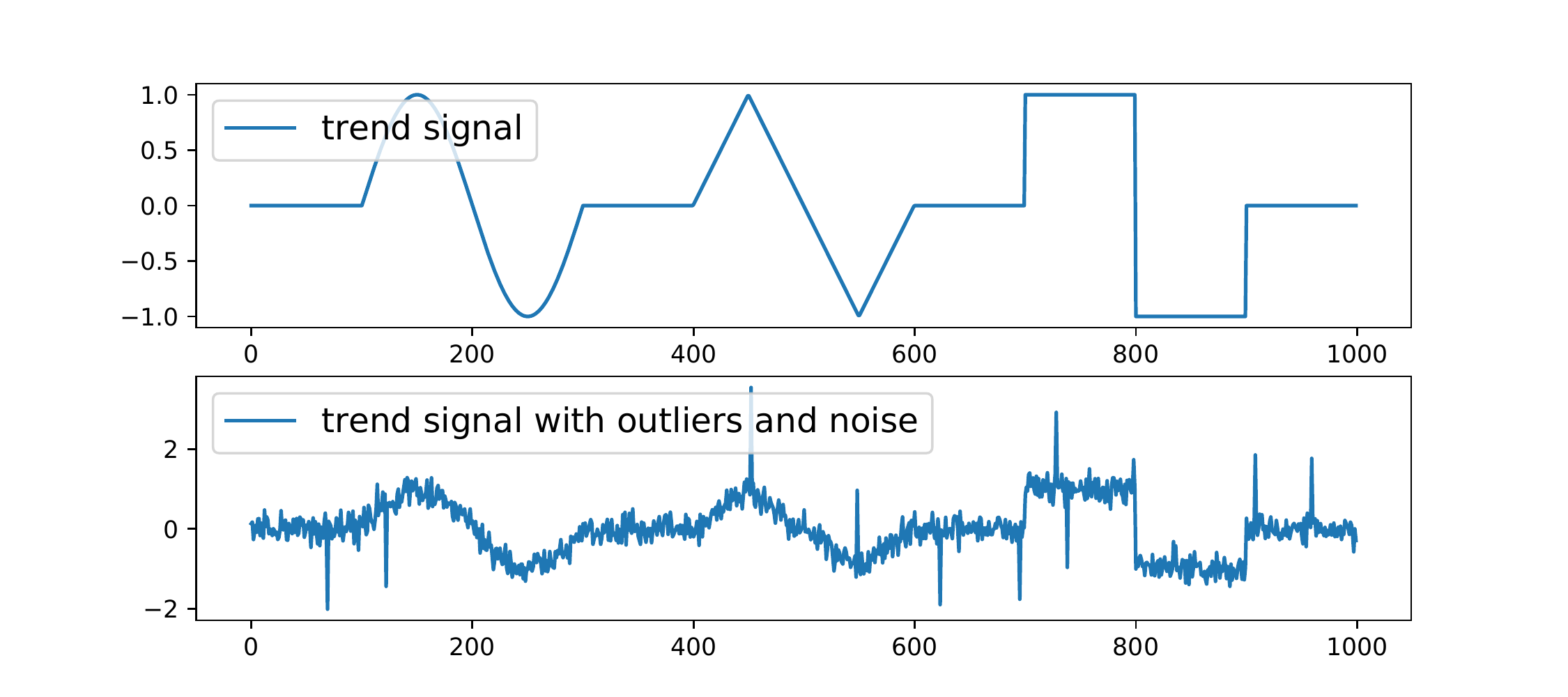} 
        \label{fig:dataset-synthetic}
        }
        \\\vspace{-0.1cm}
    \subfigure[Real-world Data]{
        \includegraphics[width=0.9\linewidth]{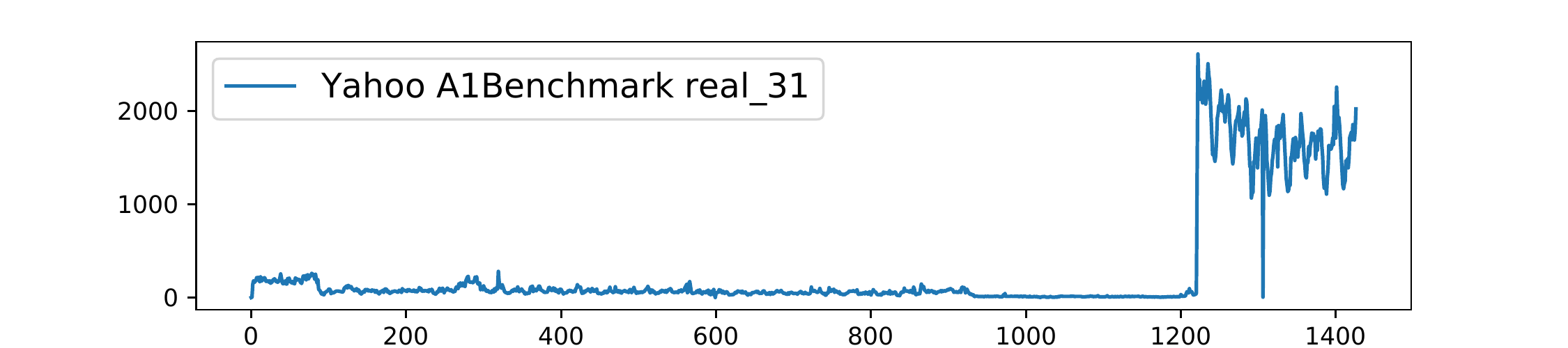}
        \label{fig:dataset-real}
        } \vspace{-0.3cm}
    \caption{Synthetic and real-world data used for the experiment.}
    \label{fig:dataset}
\end{figure}

We tested nine state-of-the-art trend filters as baseline algorithms to perform a relative comprehensive evaluation, including {H-P filter}, {$\ell_1$ trend filter}, {TV denoising filter}, {Mixed trend filter}~\cite{tibshirani2014adaptive}, {Wavelet trend filter}~\cite{craigmile2006wavelet}, {Repeated median filter}~\cite{siegel1982robust}, {robfilter}~\cite{fried2004robust}, {EMD trend filter}~\cite{cmp_l1hpemp} and its extension {EEMD trend filter}.

\subsection{Dataset}
For synthetic data, we first generate the trend signal with $1000$ data points, which contains a sin wave, a triangle wave, and a square wave with $1.0$ amplitude to represent smooth slow trend change as well as abrupt trend change with sharp discontinuity. Next, we add Gaussian noise with $0.2$ standard deviation. Then, we add $10$ to $200$ spikes and dips with $2.0$ amplitude to represent $1\%$ to $20\%$ outlier ratio. As an example, the trend signal and trend with noise and $1\%$ outlier are shown in Figure~\ref{fig:dataset-synthetic}. 

Note that we generate synthetic data with sin, triangle, and square wave as trend signal to simulate different types of trend changes in real-world data. If the trend signal only contains triangle/rectangle shape, the $\ell_1$/TV filter would be the optimal choice (under the case of white noise without outliers). However, these specific assumptions are of limited usage in practice. As the ground truth of trend is known, the mean squared error (MSE) and mean absolute error (MAE) are used for the performance evaluation later.

\begin{figure}[!h] 
    \centering
    \includegraphics[width=.82\linewidth]{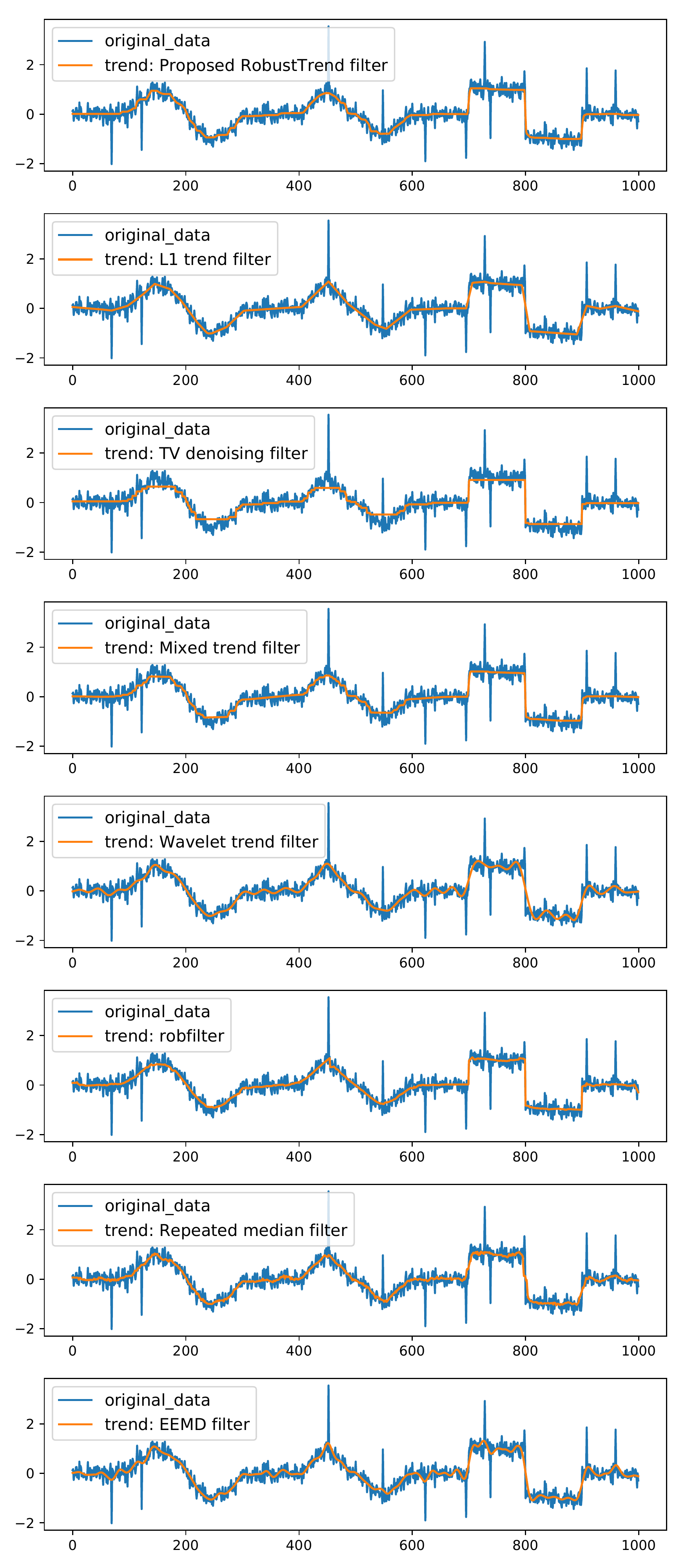}
    \caption{Trend filtering results on synthetic data.}
    \label{fig:synthetic-results}
\end{figure}

In addition to the synthetic data, we also adopt one real-world data from Yahoo's Anomaly Detection Dataset\footnote{https://webscope.sandbox.yahoo.com/catalog.php?datatype=s}, i.e., real\_31 in A1Benchmark, which is shown in Figure~\ref{fig:dataset-real}. It can be seen that there is abrupt trend change around point $1220$ and a outlier at point $1306$.

\subsection{Experiment Results on Synthetic Data}

Figure~\ref{fig:synthetic-results} shows the trend filtering results from different algorithms when there is $1\%$ outlier ratio. As can be seen in the figure, $\ell_1$ trend filter, wavelet trend filter, and EEMD trend filter are heavily affected by the abrupt level change around the square wave. TV denoising filter captures abrupt level change but exhibits staircasing around sine and triangle wave. Due to space limitation, we omit the results of H-P and EMD trend filters since their performances are worse than $\ell_1$ and EEMD trend filters, respectively.
To better evaluate the performance quantitatively, we calculate MSE and MAE to quantify the extraction accuracy. In addition to extracting trends on 1\% of outliers, we also increase the ratio of outliers to 5\%, 10\%, and 20\%. Table~\ref{tab: 1-alg-comparison-synthetic} summarizes the performance of our proposed algorithm along with other nine state-of-the-art trending filtering algorithms. The best results for each case are highlighted in bold fonts. Overall, it can be observed that our algorithm outperforms others. 

The trend recovery near change points is important as it measures how prompt we can capture the trend change in time. Thus, we calculate MSE and MAE around 9 change points and their neighbors (one before and one after each change point) in the synthetic data (total 27 points) when outlier ratio is 5\%. The results are summarized in Table~\ref{tab:changePoint-comparison-synthetic} where the best results are highlighted in bold fonts. Clearly, our algorithm achieves significantly more accurate trend near change points.

To evaluate the different components of our RobustTrend filter, we also compare TV denoising filter with Huber loss (i.e., RobustTrend filter without second order difference regularization), $\ell_1$ trend filter with Huber loss (i.e., RobustTrend filter without first order difference regularization), and RobustTrend filter with $\ell_2$-norm regularizations. The results are summarized in Table~\ref{tab:decouple-comparison-synthetic} with the outlier ratio setting to 5\%, where the best results are highlighted in bold fonts. It can be seen that the Huber loss and the first order and second order difference $\ell_1$ regularization terms bring significant performance improvements in trend filtering.

\subsection{Experiment Results on Real-World Data}

\begin{figure}[!h]
    \centering
        \includegraphics[width=0.8\linewidth]{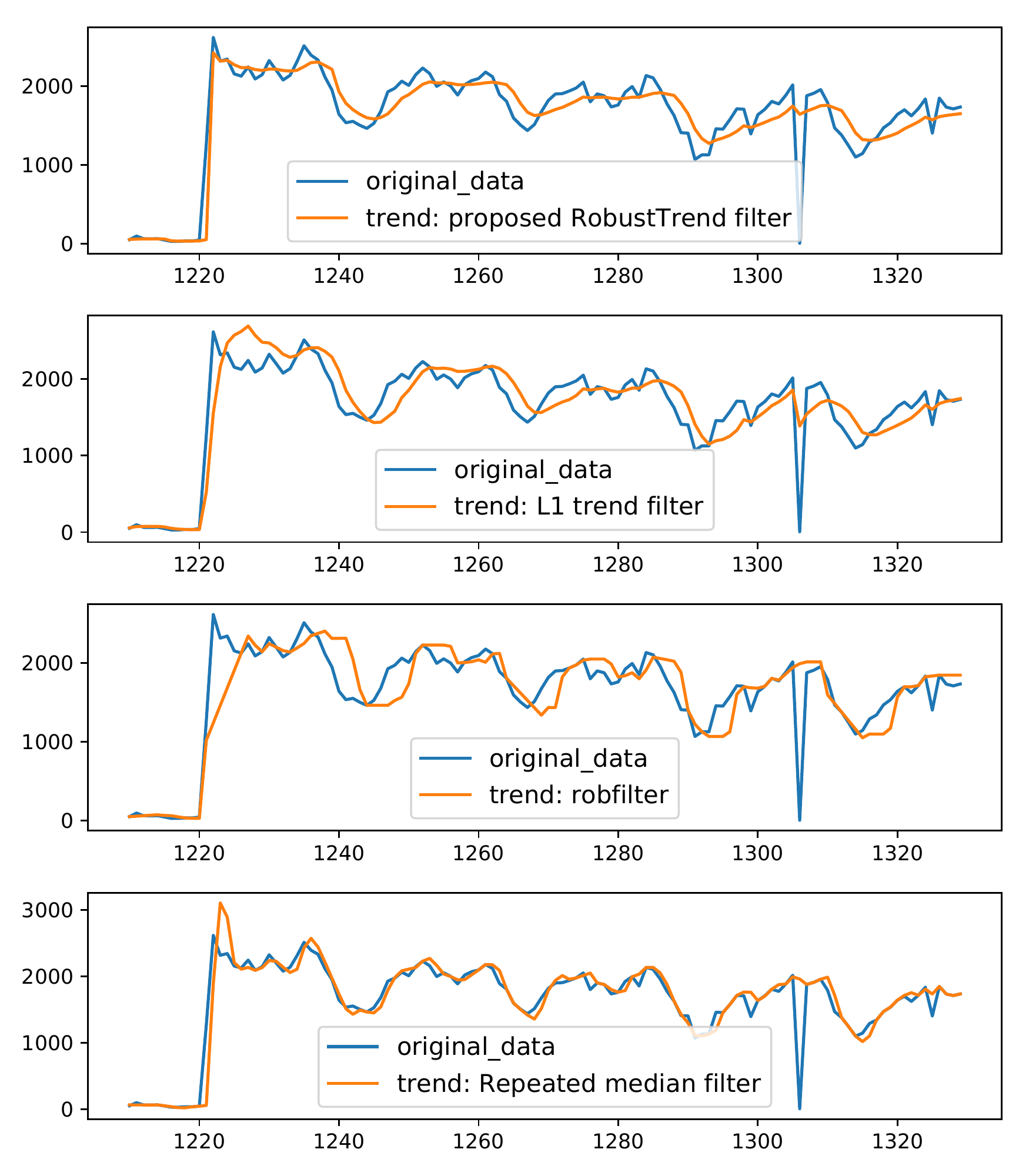}
    \vspace{-0.1cm}
    \caption{Zoomed-in version of the trend filtering results from different algorithms using online mode on real data.}
    \label{fig:real-results}
\end{figure}

We perform experiment on one real-world dataset from Yahoo depicted in Figure~\ref{fig:dataset-real}. In this experiment, we apply the online mode of the trend filtering to investigate how it performs on streaming data. The results of top-3 performance filters (our RobustTrend, robfilter, and repeated median filter) and the popular $\ell_1$ trend filter are summarized in Figure~\ref{fig:real-results}. Since the beginning of the data is almost constant, we only show the zoomed-in results after point 1200. It can be observed that our RobustTrend is able to follow the sudden increase of the trend and is not affected by the outlier. The $\ell_1$ filter and robfilter show a delay in capturing the trend change. The $\ell_1$ filter is also sensitive to the outliers in the original signal, leading to a dip in the trend. The repeated median algorithm overshoots the estimation of the trend and generates a higher trend than the actual one near the position where trend abruptly changes.

\begin{table}[!htb]
\footnotesize \centering
\begin{tabular}{l|c|c}
\hline
Algorithms&  MSE &  MAE  \\ \hline\hline
H-P filter & 0.1463 & 0.3036 \\
$\ell_1$ trend filter  & 0.1404 & 0.2910 \\
TV denoising filter  & 0.1100 & 0.2488 \\
Mixed trend filter  & 0.1140 & 0.2526 \\
Wavelet trend filter & 0.1467  & 0.3052 \\
robfilter & 0.1900 & 0.2939 \\
Repeated median filter & 0.1378 & 0.2495 \\
EMD filter & 0.1618  & 0.3074 \\
EEMD filter & 0.1570 & 0.3004 \\
Proposed RobustTrend filter & \textbf{0.0862} & \textbf{0.1966}\\\hline   
\end{tabular}
\vspace{-0.1cm}
\caption{Comparison of change points MSE and MAE on synthetic data of different trend filtering algorithms under 5\% outlier ratio.}
\label{tab:changePoint-comparison-synthetic}
\end{table}

\begin{table}[htb]
\footnotesize \centering
\begin{tabular}{l|c|c}
\hline
Algorithms&  MSE &  MAE  \\ \hline\hline
TV denoising filter & 0.0223 & 0.0983 \\
TV denoising filter with Huber loss & 0.0077 & 0.0564 \\
$\ell_1$ trend filter & 0.0170 & 0.0860 \\
$\ell_1$ trend filter with Huber loss & 0.0097 & 0.0677  \\
RobustTrend with L2 Reg & 0.0093 & 0.0585  \\
Proposed RobustTrend filter & \textbf{0.0054} & \textbf{0.0442}\\\hline   
\end{tabular}
\vspace{-0.1cm}
\caption{Comparison of MSE and MAE on synthetic data of different trend filtering algorithms under 5\% outlier ratio.}
\label{tab:decouple-comparison-synthetic}
\end{table}

%% file: 5_conclusion.tex
\section{Conclusions}



In this paper we propose a robust trend filtering method RobustTrend based on the Huber loss. The Huber loss can adaptively deal with residual outliers and different types of data. In order to impose the smoothness of the trend and meanwhile capture the abrupt trend change, we propose to use the combination of the first order and the second order difference of the trend component as regularization in trend filtering. To efficiently solve the resulting optimization problem, we design an MM-ADMM algorithm which applies majorization-minimization to facilitate the updating step for the Huber loss effectively. 



%